\documentclass[10pt]{article}
\usepackage[utf8]{inputenc}
\usepackage[margin=1in]{geometry} 
\usepackage{amsmath}
\usepackage{amsfonts}
\usepackage{amssymb}
\usepackage{graphicx}
\usepackage{cite}
\usepackage{url}
\usepackage{lineno}
\usepackage{hyperref}
\usepackage{xcolor}
\usepackage{booktabs}
\usepackage{todonotes}
\usepackage{subcaption}
\usepackage[detect-all,group-separator={,}]{siunitx}
\usepackage{multirow}
\usepackage{mdframed}
\usepackage{authblk}

\usepackage{caption}
\title{Scalable Cross-Facility Federated Learning for Scientific Foundation Models on Multiple Supercomputers}

\author[1]{Yijiang Li}
\author[2]{Zilinghan Li}
\author[3]{Kyle Chard}
\author[2,3]{Ian Foster}
\author[1]{Todd Munson}
\author[2]{Ravi Madduri}
\author[1]{Kibaek Kim}

\affil[1]{Mathematics and Computer Science Division, Argonne National Laboratory}
\affil[2]{Data Science and Learning Division, Argonne National Laboratory}
\affil[3]{Department of Computer Science, University of Chicago}

\date{}

\begin{document}

\maketitle

\begin{abstract}
Artificial Intelligence for scientific applications increasingly requires training large models on data that cannot be centralized due to privacy constraints, data sovereignty, or the sheer volume of data generated. Federated learning (FL) addresses this by enabling collaborative training without centralizing raw data, but scientific applications demand model scales that requires extensive computing resources, typically offered at High Performance Computing (HPC) facilities. Deploying FL experiments across HPC facilities introduces challenges beyond cloud or enterprise settings. We present a comprehensive cross-facility FL framework for heterogeneous HPC environments, built on Advanced Privacy-Preserving Federated Learning (APPFL) framework with Globus Compute and Transfer orchestration, and evaluate it across four U.S. Department of Energy (DOE) leadership-class supercomputers. We demonstrate that FL experiments across HPC facilities are practically achievable, characterize key sources of heterogeneity impacting the training performance, and show that algorithmic choices matter significantly under realistic HPC scheduling conditions. We validate the scientific applicability by fine-tuning a large language model on a chemistry instruction dataset, and identify scheduler-aware algorithm design as a critical open challenge for future deployments.
\end{abstract}

% \tableofcontents

\section{Introduction}
\label{sec:introduction}

Artificial Intelligence for scientific applications increasingly faces a structural mismatch between where data are generated and stored and where model training occurs. There exist two distinct barriers preventing the centralization of raw training data. The first is privacy and data sovereignty: sensitive datasets in regulated domains such as healthcare and power grid operations cannot be shared across institutional boundaries due to policy, regulation, or security requirements. The second is practical scale: experimental facilities such as synchrotron beamlines can produce terabytes of data per experiment, resulting in datasets that are too large or operationally embedded to centralize, even when sharing is technically permissible. In both cases, collaborative model training across all data sources is infeasible without moving computation to the data, a challenge particularly acute for scientific facilities, where both barriers often apply simultaneously.

Federated learning (FL) %addresses this challenge by enabling 
enables collaborative model training without moving raw data~\cite{mcmahan2017communication,kairouz2021advances}. In FL, clients train on local data and periodically send model updates to a central server, which aggregates them into a global model. There are two key distinct settings in FL: cross-device and cross-silo settings~\cite{kairouz2021advances,huang2022cross}. Cross-device FL involves many small, and often intermittently available, devices each holding small data volumes. Cross-silo FL involves a modest number of participating organizations or facilities, each usually holding large datasets and substantial compute resources, and are expected to participate reliably throughout training. FL experiments among scientific facilities naturally fit the cross-silo profile.

% Scientific cross-silo FL, however, requires compute at a scale that only HPC systems can provide. Modern scientific models, including large language models and domain foundation models, demand accelerator-scale training infrastructure. HPC facilities offer this, but introduce constraints that standard FL frameworks do not address: job schedulers with unpredictable queueing delays, strict site firewalls, heterogeneous accelerator architectures, and independent operational policies across sites. These constraints fundamentally reshape the FL system problem beyond what cloud or enterprise deployments encounter.

Cross-silo FL for large-scale scientific applications demands distributed GPU clusters with hundreds to thousands of GPU accelerators, computational resources that High Performance Computing (HPC) infrastructure is well-positioned to provide. However, HPC environments introduce constraints that standard FL frameworks do not address: job schedulers with unpredictable queueing delays, strict site firewalls, heterogeneous accelerator architectures, independent operational policies across sites, and system interruptions from both scheduled and unscheduled maintenance that can disrupt long-running workloads. These constraints fundamentally reshape the FL system problem beyond what cloud or enterprise deployments encounter.

Despite the maturity of FL and HPC as independent fields, only limited prior work has explored their integration in hybrid cloud-HPC or cross-facility settings \cite{colonnelli2024cross,li2024secure}. Prior studies mainly demonstrate feasibility, whereas our focus is a systematic empirical characterization of federated training across heterogeneous leadership-class HPC facilities. Existing cross-silo frameworks such as FATE~\cite{liu2021fate}, OpenFL~\cite{reina2021openfl}, NVIDIA FLARE~\cite{roth2022nvidia}, and FedML~\cite{he2020fedml} are primarily designed for cloud and enterprise settings and do not address HPC-specific constraints. Kim et al.~\cite{kim2024privacy} discuss privacy and computational heterogeneity challenges for cross-silo FL on HPC systems, but empirical evaluation across multiple facilities remains absent. 

% No prior work has implemented and systematically evaluated end-to-end FL training across multiple HPC facilities under realistic operational conditions.

% Here we close this gap by presenting and evaluating 
Here we present and evaluate a cross-facility FL framework designed for heterogeneous HPC environments. We evaluate this framework by using four U.S. Department of Energy (DOE) leadership-class supercomputers: Polaris and Aurora at the Argonne Leadership Computing Facility (ALCF), Frontier at the Oak Ridge Leadership Computing Facility (OLCF), and Perlmutter at the National Energy Research Scientific Computing Center (NERSC). These four systems span NVIDIA, AMD, and Intel accelerator architectures, multiple scheduler types, and distinct security and scheduling policies, making them a strong proxy for the diversity of modern HPC infrastructure. Our framework is built on the Advanced Privacy-Preserving Federated Learning framework (APPFL)~\cite{ryu2022appfl,li2025advances} with orchestration via Globus Compute~\cite{chard2020funcx} for training task dispatch and Globus Transfer~\cite{10943420241281744} for model and training configuration transfer, enabling local training jobs to execute under site-specific scheduling and security policies while model updates are exchanged reliably across wide-area networks.

Our contributions are as follows:
\begin{enumerate}
\item \textbf{Cross-HPC facility FL framework}: We design and implement a generalizable FL framework for orchestrating training across heterogeneous HPC facilities, supporting diverse models, datasets, and scientific tasks, while addressing challenges in communication, scheduling, and computation unique to this setting.
\item \textbf{Performance characterization}: We systematically characterize heterogeneity in computational throughput, communication costs, and queueing dynamics, revealing performance variations that existing FL algorithms do not adequately address.
\item \textbf{Algorithm evaluation}: We evaluate various existing FL algorithms under large-scale synchronous co-scheduled conditions and small-scale runs under realistic queueing conditions, revealing how heterogeneous schedulers and memory constraints impact algorithm performance.
\item \textbf{Scientific validation}: We verify the framework's applicability to real scientific workloads by fine-tuning a large language model (LLM) on a chemistry instruction dataset, demonstrating that cross-HPC facility FL can support large-scale scientific model development.
\end{enumerate}

\section{Results}
\label{sec:results}

In this section, we present experimental results for cross-HPC facility FL and provide the empirical groundings that inform the design principles of cross-HPC FL frameworks discussed in Section~\ref{sec:discussions}. In the experiment, we evaluate our framework through federated fine-tuning of an LLM on a chemistry instruction dataset. We distribute this task across four DOE leadership-class supercomputers acting as clients, with a CPU-only cluster at Argonne National Laboratory serving as the central server. We conduct two sets of experiments: (1) large-scale co-scheduled runs using HPC reservations to characterize upper-bound performance of cross-facility FL, and (2) smaller queue-based runs to evaluate algorithmic robustness under realistic scheduling conditions. Full details of hardware, software, and orchestration are provided in Section~\ref{sec:methods}.

\subsection{Use Case and Data Distribution}
We use SMolInstruct~\cite{yu2024llasmol}, a chemistry instruction tuning dataset containing over three million training samples, as our use case. SMolInstruct organizes its data into 14 distinct \textit{tasks}, each corresponding to a specific chemistry problem type, which are further grouped into four broader \textit{task groups}: name conversion, property prediction, chemical reaction, and molecule description. The number of \textit{samples} (i.e., instruction--response pairs) varies substantially across tasks, ranging from roughly 1,100 samples for the task of ESOL (Water Solubility) to nearly 978,000 for the task of forward synthesis. We fine-tune a pretrained Llama2-7B model~\cite{touvron2023llama2openfoundation} on this dataset in a federated manner.

In our FL setup, we use four leadership-class supercomputers, Aurora, Frontier, Perlmutter, and Polaris, each as an FL \textit{client}, performing local model training on its assigned data. We use a CPU-only cluster at Argonne National Laboratory, Improv, as the central \textit{server}, coordinating global model aggregation. 

We distribute samples to clients by the \textit{task groups}: Perlmutter handles samples from the four name conversion tasks, Polaris manages samples from the six property prediction tasks, Aurora focuses on samples from the two chemical reaction tasks, and Frontier covers samples from the two molecule description tasks. This partitioning creates a natural non-IID data distribution across clients, mirroring real-world scenarios where different institutions possess specialized knowledge and proprietary data in specific subdomains, such as pharmaceutical companies with drug discovery data, research labs with synthesis expertise, and academic centers with molecular property datasets. The details of the instruction tuning dataset, SMolInstruct, and its distribution to the supercomputers are presented in Table~\ref{tab:data_distribution}.

\begin{table}[!htbp]
\centering
\small 
\caption{We distribute the 3.3M SMolInstruct training data samples across four facilities and group them by task group as follows. Refer to Yu et al.~\cite{yu2024llasmol} for more details on the tasks and samples.}
\label{tab:data_distribution}
\begin{tabular}{lllll}
\toprule
&  &  & \multicolumn{2}{c}{\textbf{Samples}}  \\
\textbf{Task Group} & \textbf{HPC} & \textbf{Task} & \textbf{Per Task} & \textbf{Per Group} \\
\midrule
\multirow{4}{*}{Name Conversion} & \multirow{4}{*}{Perlmutter} & IUPAC to molecular formula & 304,490 & \multirow{4}{*}{1,217,627}\\
& & IUPAC to SMILES & 304,379 &\\
& & SMILES to molecular formula & 304,379 &\\
& & SMILES to IUPAC & 304,379 & \\
\midrule 
\multirow{6}{*}{Property Prediction} & \multirow{6}{*}{Polaris} & ESOL(Water Solubility) & 1,111 &  \multirow{6}{*}{78,319}\\
& & Lipo (Octanol/Water Distribution Coefficient) & 4,200 \\
& & BBBP (Blood-Brain Barrier Penetratio) & 1,962 \\
& & ClinTox (Toxicity to Human Body) & 1,431 \\
& & HIV (HIV Replication Inhibition) & 41,075 \\
& & SIDER (Side Effects of Drugs) & 28,540 &  \\
\midrule
\multirow{2}{*}{Molecule Description} & \multirow{2}{*}{Frontier} & Molecule captioning & 60,305 & \multirow{2}{*}{120,565}\\
 & & Molecule generation & 60,260 &  \\
\midrule
\multirow{2}{*}{Chemical Reaction} & \multirow{2}{*}{Aurora} & Forward synthesis & 977,920 & \multirow{2}{*}{1,925,903}\\
 & & Retrosynthesis & 947,983 & \\
\midrule
\textbf{Overall} & & & & \textbf{3,342,414} \\
\bottomrule
\end{tabular}
\end{table}

\subsection{FL Configurations}
We conduct two sets of experiments to evaluate our framework under different operational conditions. First, we utilize HPC reservations of 64 nodes at each of the four supercomputers to enable co-scheduling, where all systems execute synchronously, a common approach in distributed computing~\cite{czajkowski2002snap}. This approach eliminates unpredictable queueing delays and supports large-scale deployment. Under these co-scheduled conditions, we focus on a single synchronous FL algorithm (FedAvg~\cite{mcmahan2017communication}) because the absence of queueing uncertainty allows us to tune local training parameters based on preliminary computational capability estimates at each supercomputer, effectively eliminating stragglers and ensuring synchronized global aggregation. This controlled configuration isolates the performance characteristics of FL at scale from the effects of scheduling variability, enabling us to demonstrate and assess the upper bound of cross-facility FL performance when computational heterogeneity can be proactively managed. 

Second, we conduct experiments without reservations, where each client job enters the standard scheduling queue before running local training. For this scenario, we deploy a smaller configuration of two nodes per supercomputer, which is more accessible in the presence of queueing delays and scheduling randomness. This setup allows us to systematically benchmark four FL algorithms, FedAvg~\cite{mcmahan2017communication}, FedAsync~\cite{xie2019asynchronousfederatedoptimization}, FedBuff~\cite{nguyen2022fedbufffederatedlearning}, and FedCompass~\cite{li2024fedcompass},  under heterogeneous queueing policies, scheduling behaviors, and computational capabilities across facilities. These algorithms represent different synchronization paradigms in FL: FedAvg follows a fully synchronous aggregation scheme, FedAsync and FedBuff adopt asynchronous updates, and FedCompass leverages a semi-asynchronous design using a computational power-aware scheduler to balance staleness tolerance and coordination. This dual experimental design allows us to assess both the optimized performance potential of cross-facility FL under co-scheduled conditions and its algorithmic robustness under real-world HPC constraints.

\subsection{Scalability and Throughput}
\label{sec:scaling}
We evaluate throughput performance across the four supercomputers with configurations detailed in Table~\ref{tab:throughput_config}. Throughput is measured as average samples processed per second during one-hour runs, with experiments scaling from one to 64 nodes. To ensure a fair comparison of raw computational capability, all supercomputers train on the same dataset rather than their assigned task partitions, isolating throughput differences to hardware and configuration rather than data distribution.

For each supercomputer, we follow the recommended optimization flags and system-specific configurations provided by the operating facilities, meaning different supercomputers employ different settings. This is intentional, as we are comparing the best achievable performance at each site rather than enforcing a uniform configuration. Polaris and Frontier additionally leverage AWS OFI NCCL/RCCL~\cite{aws-ofi-nccl} plugins that enable high-performance collective communication over fabric interfaces such as Slingshot.

\begin{figure}[!ht]
    \centering
    \includegraphics[width=1.0\linewidth]{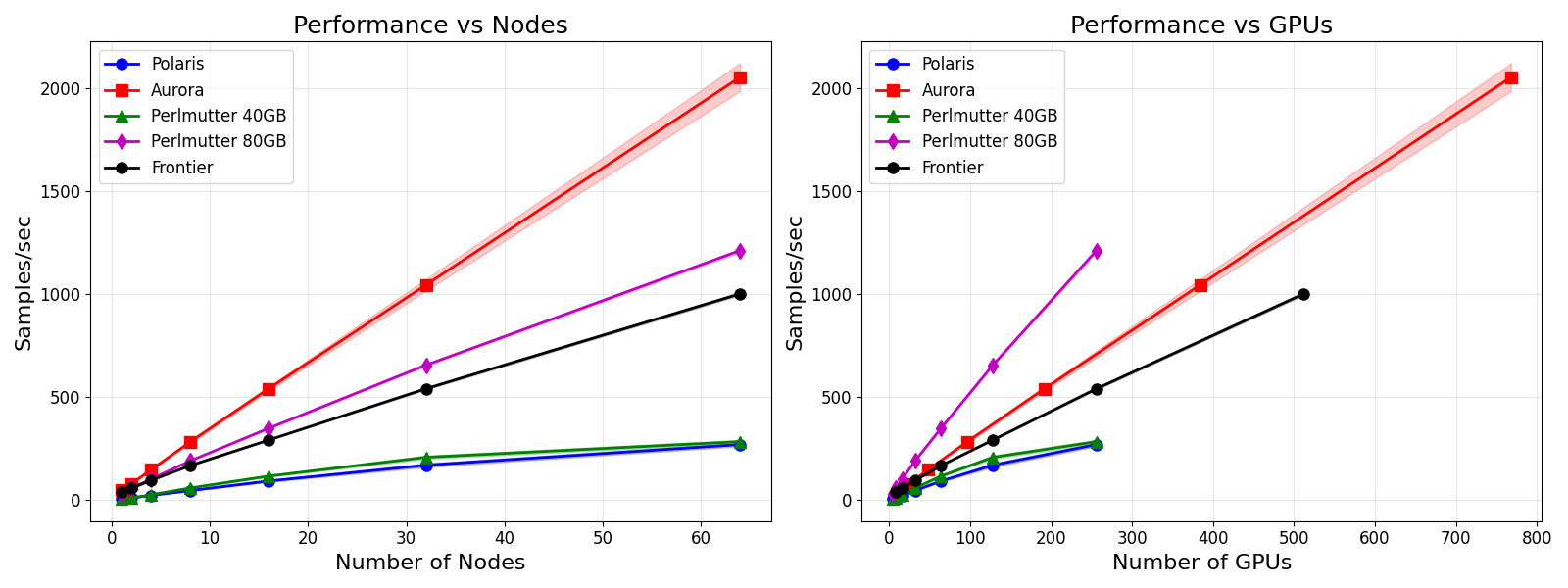}
    \caption{Throughput scaling with fixed micro-batch size per GPU. The left panel shows throughput (samples per second) as a function of node count, while the right panel shows the same results as a function of total GPU count. The consistent scaling patterns across both panels confirm that memory capacity and the resulting optimization strategy, rather than node or GPU count alone, are the primary drivers of throughput heterogeneity across supercomputers.}
    \label{fig:scaling_plot}
\end{figure}

Figure~\ref{fig:scaling_plot} reveals dramatic heterogeneity in both absolute throughput and scaling efficiency. Aurora achieves the highest throughput, reaching over 2,100 samples per second at 64 nodes with nearly linear scaling throughout. Perlmutter 80GB and Frontier %demonstrate strong intermediate performance at 
achieve approximately 1,200 and 1,000 samples per second respectively at maximum scale, while Polaris and Perlmutter 40GB both plateau around 250 samples per second.

The performance disparity reflects a fundamental trade-off driven by GPU memory capacity. Smaller memory GPUs require aggressive memory optimization strategies such as DeepSpeed ZeRO-3~\cite{rasley2020deepspeed, rajbhandari2020zero}, which achieve memory efficiency at the cost of increased communication overhead. Larger memory GPUs avoid this penalty by using less communication-intensive approaches such as ZeRO-1. This is most clearly illustrated by Perlmutter's two configurations: the 40GB variant must employ ZeRO-3 and achieves only 250 samples per second, while the 80GB variant uses ZeRO-1 and reaches 1,200 samples per second, an over four-fold difference attributable entirely to the memory-driven optimization trade-off.

When viewed by total GPU count rather than node count, these scaling patterns remain consistent, confirming that memory capacity and the resulting optimization strategy are the primary drivers of observed heterogeneity. This variation, from 250 samples per second for memory-constrained configurations to over 2,100 for Aurora, underscores a critical challenge for cross-facility FL: facilities exhibit qualitatively different scaling behaviors that make computational capability-aware algorithms like FedCompass essential. Naive approaches that treat all clients equally result in severe load imbalance, with faster clients waiting on slower clients in synchronous settings, or slower clients' contributions being overwhelmed in asynchronous settings.

\begin{table}[!htbp]
\centering
\caption{Configurations for throughput scaling experiments. GPU memory capacity largely determines the required DeepSpeed ZeRO optimization stage: supercomputers with smaller memory GPUs must employ ZeRO-3, which achieves memory efficiency at the cost of increased inter-GPU communication overhead, while supercomputers with larger memory GPUs can use the less communication-intensive ZeRO-1. Micro batch size is set per GPU and held fixed across node counts.}
\begin{tabular}{llccc}
\toprule
\textbf{Supercomputer} & \textbf{GPU architecture} & \textbf{GPU memory} & \textbf{DeepSpeed stage} & \textbf{Micro batch size} \\
\midrule
Polaris & NVIDIA A100 & 40GB & ZeRO-3 & 6\\
Perlmutter & NVIDIA A100 & 40GB & ZeRO-3 & 6\\
Perlmutter & NVIDIA A100 & 80GB & ZeRO-1 & 16\\
Aurora & Intel Max 1550 Series & 64GB & ZeRO-1 & 8\\
Frontier & AMD MI250X & 64GB & ZeRO-1 & 12\\
\bottomrule
\end{tabular}
\label{tab:throughput_config}
\end{table}

\subsection{Training Performance}

We evaluate the training performance under the two experiment configurations mentioned previously: large-scale co-scheduled deployment (64 nodes per supercomputer starting synchronously using reservations) and small-scale queued deployment (two nodes per supercomputer with algorithm comparison). All experiments evaluate models on all 14 tasks across the dataset, even for local client models, to demonstrate knowledge transfer from clients to server without requiring data movement. Our experiments focus on demonstrating a functional cross-facility FL framework rather than training to full convergence; accordingly, we conduct eight global (synchronous) rounds for the 64-node experiment and distribute 40 total local training rounds across clients for the two-node comparison. Model evaluation is performed on a fixed subset of the test set using standard evaluation protocols with deterministic random seeds to ensure fair comparison across experiments.

\subsubsection{Large-Scale Co-Scheduled Deployment}

\begin{figure}[!htbp]
    \centering
    \includegraphics[width=0.7\linewidth]{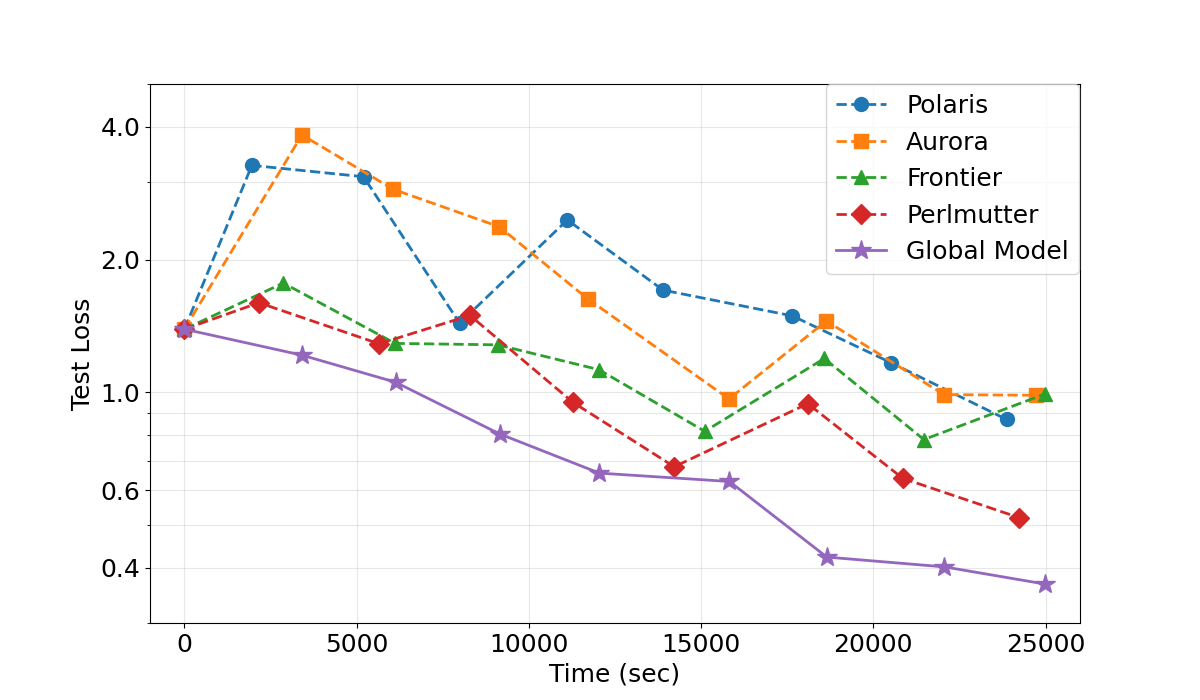}
    \caption{FedAvg test loss over eight global aggregation rounds using 64 nodes per supercomputer (client). The global model (purple) consistently achieves the lowest test loss across all 14 tasks, decreasing from 1.39 to 0.37, while local client models show higher losses reflecting specialization to their assigned task groups.}
    \label{fig:fedavg_64nodes}
\end{figure}

Figure~\ref{fig:fedavg_64nodes} presents the training dynamics observed for FedAvg across all four supercomputers under co-scheduled conditions with 64 nodes per supercomputer, utilizing over 1,700 GPUs in total.  Hardware specifications are shown in Table~\ref{tab:hardware_specs_64nodes}. We tune local training steps at each supercomputer based on throughput measurements from our scaling study in Section~\ref{sec:scaling}, targeting approximately 40 minutes per local training round including initialization and computational overhead. This tuning ensures near-simultaneous completion across the supercomputers.

This experiment demonstrates the fundamental functioning of FL at scale, with the global model loss decreasing consistently and monotonically from 1.39 to 0.37 over eight aggregation rounds, confirming that the framework successfully orchestrates collaborative learning across distributed supercomputers. 
The results inform four general observations. 

First, the global model, represented by the purple line, consistently achieves the lowest test loss throughout training, demonstrating that the federated aggregation successfully integrates knowledge across the distributed, domain-specialized datasets. This validates that our framework enables effective cross-facility collaborative learning without raw data centralization.

Second, all individual client models exhibit higher test loss when evaluated on all tasks compared to the global model, with notable variance across clients. This behavior reflects the client drift phenomenon~\cite{karimireddy2020scaffold, li2020convergencefedavgnoniiddata}: as each client trains exclusively on its specialized domain subset (with 64 nodes enabling more samples per local training round), local models optimize toward domain-specific patterns that may not generalize well to the full task distribution. Polaris and Aurora, handling property prediction and chemical reaction tasks respectively, show particularly high losses on the full evaluation set, underscoring their specialization. 

Third, the consistent performance ordering across supercomputers reveals interesting patterns about task transferability. Perlmutter, despite handling only name conversion tasks (conversions between IUPAC, SMILES, and Molecular Formula), achieves the lowest test losses on the full evaluation. Name conversion tasks require broad knowledge of molecular representations that underlies much of chemistry, which naturally limits local model drift. Conversely, Aurora exhibits higher losses from chemical reaction tasks, indicating that synthesis and retrosynthesis reasoning is highly specialized and less generalizable. Polaris, with the smallest total sample count distributed across six diverse property prediction tasks (solubility, toxicity, blood-brain barrier permeability, HIV inhibition, ClinTox, and SIDER), shows the highest losses, reflecting the effect of limited samples available for this task group. This ordering persists across training, suggesting fundamental differences in how knowledge from different chemistry domains contributes to the federated model. 

Fourth, despite different computational characteristics and task distributions, all clients demonstrate consistent downward loss trajectories, with the global model maintaining steady improvement. The synchronized nature of co-scheduled training is evident in the smooth, regular updates across all clients. 

\begin{table}[ht]
\centering
\caption{Hardware specifications for the 64-node large-scale co-scheduled experiment. Aurora and Frontier's higher total GPU counts result in substantially larger effective 
batch sizes compared to Polaris and Perlmutter. Polaris operated with 63 nodes due to a single node failure during the reservation.}
\begin{tabular}{llccc}
\toprule
\textbf{Facility} & \textbf{GPU architecture} & \textbf{Nodes} & \textbf{Total GPUs} & \textbf{Effective batch size} \\
\midrule
Polaris & NVIDIA A100 & 63 (1 node failure) & 252 & \num{1512} \\
Perlmutter & NVIDIA A100 40GB & 64 & 256 & \num{1536} \\
Aurora & Intel Max 1550 Series & 64 & 768 & \num{6144} \\
Frontier & AMD MI250X & 64 & 512 & \num{4096} \\
\bottomrule
\end{tabular}
\label{tab:hardware_specs_64nodes}
\end{table}

\subsubsection{Small-Scale Algorithm Comparison}

We next examine results from two-node experiments without HPC reservations, comparing FedAvg, FedAsync, FedBuff, and FedCompass, under heterogeneous queueing conditions, where each local training job enters the standard scheduling queue of the corresponding HPC facility. Table~\ref{tab:hardware_specs_2nodes} details the hardware specifications, showing the same per-GPU batch size as the large-scale co-scheduled deployment. The four algorithms handle heterogeneity in training completion times differently: FedAvg must wait for all clients to complete each round before aggregation, making it sensitive to stragglers. FedAsync allows clients to update the global model independently, tolerating delays at the clients. FedBuff aggregates updates in batches, offering a middle ground between synchrony and flexibility. FedCompass adapts local training steps to individual computational throughput, enabling more balanced contributions across heterogeneous clients.  All algorithms perform 40 total local training rounds distributed across clients, with initial local training steps set proportional to sample size in the assigned task groups, creating inherent heterogeneity in training completion times. Table~\ref{tab:training_distribution} shows the resulting distribution of training rounds per client: FedAvg distributes rounds evenly due to its synchronous aggregation requirement, while asynchronous algorithms produce uneven distributions driven by each client's computational speed and queue behavior, with FedCompass partially counteracting this imbalance through its adaptive step adjustment.

\begin{figure}[!htbp]
    \centering
    \includegraphics[width=1.0\linewidth]{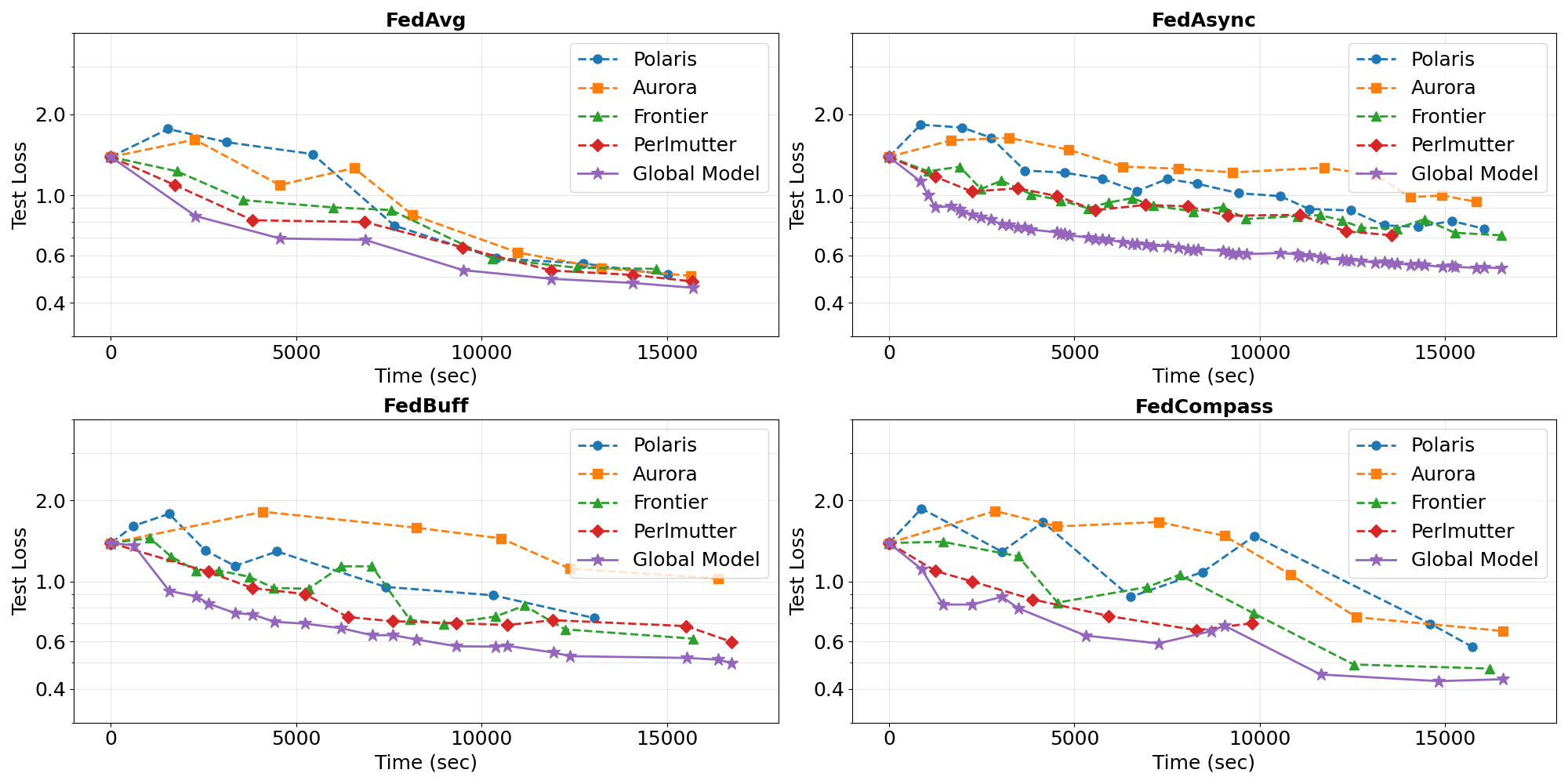}
    \caption{Test loss progression of four FL algorithms under realistic queueing conditions using two nodes per facility. Each panel shows local client models alongside the global aggregated model within the same wall-clock time budget of 17,000 seconds. FedCompass achieves the lowest final global test loss, while FedAvg produces the most stable client trajectories.}
    \label{fig:fed_algos}
\end{figure}

Figure~\ref{fig:fed_algos} compares the loss progression of the local models at each client and the aggregated global model from the four algorithms. To ensure a fair comparison, all algorithms are evaluated within the same wall-clock time budget of 17,000 seconds, allowing us to assess convergence efficiency under real-world uncertain queue conditions. In each panel, a data point for a client model appears when that client completes a local training round, while a data point for the global model marks a global aggregation event. The four algorithms exhibit distinct behaviors under these conditions.

In particular, within the given wall-clock time budget, FedCompass achieves the best global model performance with a final test loss of 0.4345, representing improvements of 4.5\% upon FedAvg, 19.3\% upon FedAsync, and 12.9\% upon FedBuff, respectively. FedBuff outperforms FedAsync by moderating the impact of stragglers through buffered aggregation. Across all algorithms, the global model consistently outperforms individual local models on the full task evaluation test set, reinforcing that FL successfully transfers knowledge across domain-specialized datasets.

Examining client trajectories reveals further distinctions. FedAvg shows the most consistent client convergence, with all clients following smooth downward trajectories that closely track the global model. Individual local models maintain similar loss values throughout training with minimal divergence. FedAsync and FedBuff produce notably erratic client behavior, particularly for Polaris and Aurora, whose loss trajectories oscillate significantly even as the global model improves steadily. FedCompass displays the most volatile individual client behavior, with pronounced oscillations across clients, due to its adaptive step adjustments. Despite this volatility, FedCompass' global model achieves the largest overall improvement, ultimately reaching the lowest final loss among all algorithms.

These results highlight a key insight: algorithms that explicitly account for computational heterogeneity outperform those that merely tolerate asynchrony. FedAsync and FedBuff reduce the impact of stragglers but treat all clients as roughly equivalent contributors, which can bias the global model toward faster clients' data distributions. FedCompass' superior performance demonstrates that proactively adapting to known computational differences yields better outcomes. However, FedCompass does not yet model queue time variability or system-specific scheduling policies, suggesting a clear avenue for further improvement.

\begin{table}[ht]
\centering
\caption{Hardware specifications for the two-node small-scale experiment. Same as the 64-node setup, Aurora and Frontier's higher multi-GPU-per-node architectures result in larger effective batch sizes relative to Polaris and Perlmutter despite the same node count.}
\begin{tabular}{llccc}
\toprule
\textbf{Facility} & \textbf{GPU architecture} & \textbf{Nodes} & \textbf{Total GPUs} & \textbf{Effective batch size} \\
\midrule
Polaris & NVIDIA A100 & 2 & 8 & \num{48} \\
Perlmutter & NVIDIA A100 40GB & 2 & 8 & \num{48} \\
Aurora & Intel Max 1550 Series & 2 & 24 & \num{192} \\
Frontier & AMD MI250X & 2 & 16 & \num{128} \\
\bottomrule
\end{tabular}
\label{tab:hardware_specs_2nodes}
\end{table}

\begin{table}[!htbp]
\centering
\caption{Distribution of local training rounds across four clients for each algorithm over 40 total rounds. FedAvg distributes rounds evenly due to its synchronous aggregation requirement. Asynchronous algorithms produce uneven distributions driven by each client's computational speed and queue behavior.}
\label{tab:training_distribution}
\begin{tabular}{lccccc}
\toprule
\textbf{Algorithm} & \textbf{Total} & \textbf{Polaris} & \textbf{Aurora} & \textbf{Frontier} & \textbf{Perlmutter}  \\
\midrule
FedAvg & 40 & 10 & 10 & 10 & 10 \\
FedAsync & 40 & 11 & 6 & 14 & 9 \\
FedCompass & 40 &  11 & 9 & 11 & 9 \\
FedBuff & 40 & 8 & 5 & 17 & 10\\
\bottomrule
\end{tabular}
\end{table}

\subsection{Globus Transfer Communication}

When using FL to train large models with billions of parameters, communication overhead becomes an increasingly important factor and can emerge as a significant performance bottleneck, particularly in cross-facility deployments where large model parameters must be transferred across geographically distributed HPC facilities. To quantify this cost, we analyze the efficiency of transferring LLMs across clients using Globus Transfer, with all transfers originating from the server to the four clients. We evaluate five models ranging from OPT-125m to Llama2-13b in BF16 format. For a comprehensive review of Globus Transfer and its performance characteristics, we refer readers to~\cite{10943420241281744}.

\begin{figure}
    \centering
    \includegraphics[width=1\linewidth]{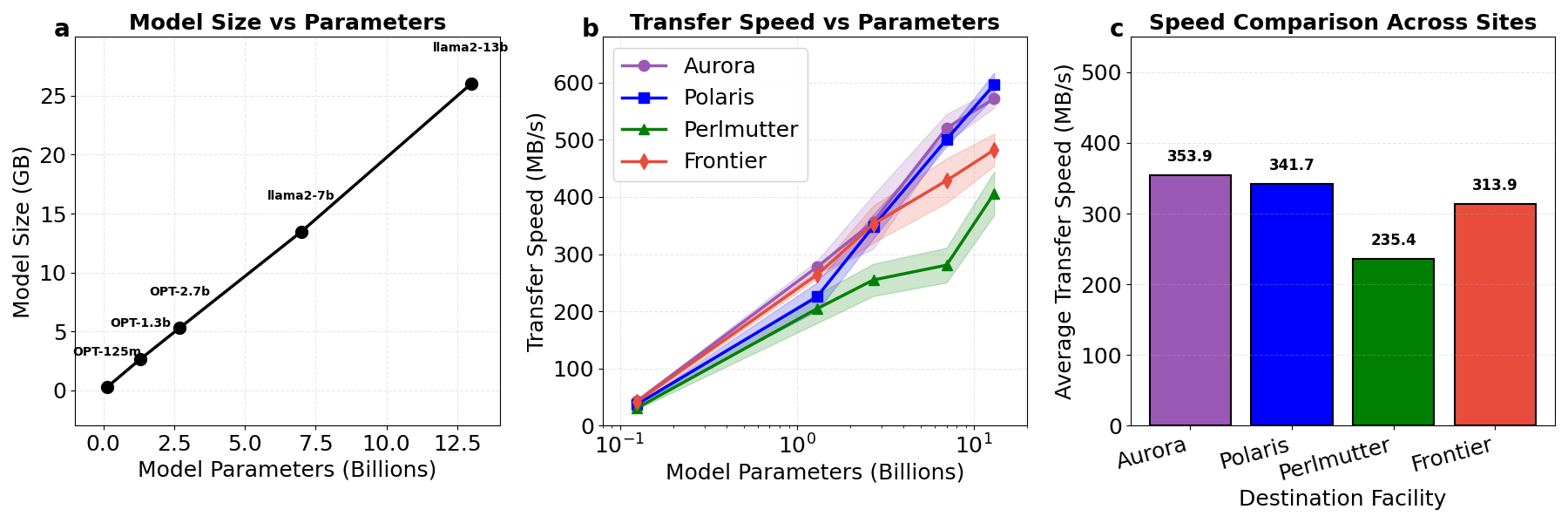}
    \caption{Globus Transfer communication efficiency across clients for five models ranging from OPT-125m to Llama2-13b. Panel \textbf{a} shows the linear relationship between model parameters and storage size in BF16 format. Panel \textbf{b} shows transfer speed versus model size, with Aurora and Polaris achieving the highest speeds due to co-location with the server at Argonne. Panel \textbf{c} summarizes average transfer speeds over all five models per destination facility.}
    \label{fig:communication_efficiency}
\end{figure}

Figure~\ref{fig:communication_efficiency} presents three views of transfer performance. Panel \textbf{a} demonstrates the linear relationship between model parameters and storage requirements, scaling from approximately 250 MB for 125 million parameters to 26 GB for 13 billion parameters. Panels \textbf{b} and \textbf{c} reveal how transfer speeds vary significantly across both clients and model sizes.

Model size significantly impacts transfer efficiency, with speeds increasing substantially from the smallest to the largest models, indicating that larger transfers better amortize connection establishment overhead and benefit more from high-bandwidth local networks. Aurora and Polaris demonstrate the strongest scaling behavior, improving more than 10-fold in transfer speed from OPT-125m to Llama2-13b, while Perlmutter exhibits slightly weaker scaling. 

Aurora and Polaris consistently achieve the highest transfer speeds, which we attribute to co-location: both the server and ALCF destinations (hosting Aurora and Polaris) reside at Argonne National Laboratory, enabling transfers over high-bandwidth local networks. Table~\ref{tab:latency} confirms this co-location advantage quantitatively: Aurora and Polaris exhibit sub-millisecond round-trip latency to the server, compared to 17ms for Frontier and 45ms for Perlmutter, directly reflecting the difference between intra-campus and wide-area network paths.

\begin{table}[!htbp]
\centering
\caption{Round-trip latency (ms) between the server and each client, measured via ping over 20 rounds. Aurora and Polaris exhibit 
sub-millisecond latency due to co-location with the server at Argonne National Laboratory, while Frontier and Perlmutter face substantially higher wide-area network latency.}
\begin{tabular}{lcccc}
\toprule
\textbf{Destination} & \textbf{Min} & \textbf{Avg} & \textbf{Max} & \textbf{Mean Dev.} \\
\midrule
Aurora (ALCF)      & 0.122 & 0.266 & 0.432 & 0.094 \\
Polaris (ALCF)     & 0.112 & 0.210 & 0.342 & 0.074 \\
Frontier (OLCF)    & 17.236 & 17.281 & 17.321 & 0.167 \\
Perlmutter (NERSC) & 45.178 & 45.205 & 45.264 & 0.286 \\
\bottomrule
\end{tabular}
\label{tab:latency}
\end{table}

Notably, the performance gap between clients widens substantially with model size. For OPT-125m, speeds are comparable across all destinations, but for Llama2-13b the gap expands to nearly 200 MB/s. The weaker scaling at Perlmutter suggests that some wide-area network paths may be bandwidth-limited, preventing large transfers from fully exploiting available link capacity.

These findings have important implications for cross-facility FL and complement recent cross-silo communication-compression efforts \cite{zhang2025fedcspc}. Transfer costs vary significantly by client location and model size, and speeds can be further influenced by temporal factors such as facility workload, time of day, and network congestion, introducing variability that FL algorithms must accommodate. In practical deployments, these heterogeneous transfer characteristics necessitate careful consideration of FL scheduling and aggregation strategies.

\section{Discussion}
\label{sec:discussions}

We discuss related work and broader context situating our contributions, and two aspects informed by our results: practical considerations for future FL algorithm design on HPC infrastructure, and lessons learned from our cross-facility deployments.

\subsection{Related Work and Context}
FL has matured across two broad deployment contexts~\cite{kairouz2021advances}. \textit{Cross-device} FL, exemplified by Google's production systems~\cite{bonawitz2019federatedlearningscaledesign}, Meta's Papaya~\cite{huba2022papayapracticalprivatescalable}, and Apple's on-device personalization~\cite{paulik2021federatedevaluationtuningondevice}, targets large populations of resource-constrained devices. \textit{Cross-silo} FL connects institutional participants with substantial compute, supported by frameworks including FATE~\cite{liu2021fate}, OpenFL~\cite{reina2021openfl}, NVIDIA FLARE~\cite{roth2022nvidia}, FederatedScope~\cite{federatedscope}, Flower~\cite{beutel2020flower}, FedML~\cite{he2020fedml}, and Flight \cite{hudson2025flight}. Our work employs APPFL \cite{ryu2022appfl,li2025advances}, an open-source privacy-preserving FL framework designed for scalability across heterogeneous computing environments, which we extend to operate across geographically distributed HPC facilities.

FL algorithm development has addressed heterogeneity from multiple angles. FedProx~\cite{li2020federatedoptimization} and SCAFFOLD~\cite{karimireddy2020scaffold} tackle statistical heterogeneity under non-IID data distributions. Asynchronous approaches including FedAsync~\cite{xie2019asynchronousfederatedoptimization} and FedBuff~\cite{nguyen2022fedbufffederatedlearning} reduce the impact of stragglers, while FedCompass~\cite{li2024fedcompass} adapts local training steps to computational throughput. Hierarchical FL~\cite{liu2020clientedge} and related multi-tier systems~\cite{chai2021fedat} organize aggregation to exploit network topology and mitigate heterogeneity. Adaptive server-side optimizers such as FedYogi and FedAdam~\cite{reddi2020adaptive} improve convergence under heterogeneous client datasets. However, none of these algorithms model batch job scheduling dynamics or wide-area communication variability, which are shown to be crucial factors in cross-HPC FL performance from our results.

Modern HPC facilities rely on job schedulers, such as SLURM~\cite{yoo2003slurm}, PBS~\cite{10.1007/3-540-60153-8_34}, and Cobalt~\cite{cobalt2010}, to manage competing workloads through priority queues and backfilling.  These schedulers are often designed for independent batch jobs rather than iterative federated workloads. Multi-site workflow systems including Condor~\cite{frey2002condor}, Pegasus~\cite{deelman2015pegasus}, and StreamFlow~\cite{colonnelli2024cross} coordinate distributed scientific computations but lack FL's iterative synchronization requirements. Predictive scheduling~\cite{gainaru2019schedulingscientificworkflows} and performance variability analysis~\cite{hoefler2017performance} inform the understanding of HPC dynamics but have not been applied to FL. Large-scale distributed training frameworks including DeepSpeed with ZeRO optimization and communication primitives from Horovod~\cite{sergeev2018horovod} provide the foundation for local training at each facility, with gradient compression techniques such as PowerSGD~\cite{chen2018powersgd} further relevant for bandwidth-constrained wide-area settings.

FL is increasingly applied across scientific domains where data sensitivity or sovereignty prevents centralization. Medical imaging~\cite{rieke2020future,sheller2020federated} and pharmaceutical research through MELLODDY~\cite{melloddy2022collaborative} demonstrate collaborative model development across institutions. Power grid applications present a particularly compelling case: foundation models for electrical load prediction and AC optimal power flow have demonstrated strong performance~\cite{bose2024rnnsfoundationmodelsempirical,li2026luminafoundationmodelstopology}, and federated training enables collaboration without sharing operational data~\cite{bose2023privacypreservingloadforecastingpersonalized}. Light source facilities face similar challenges, generating massive imaging datasets where competing scientific interests necessitate federated approaches. Model pruning techniques~\cite{bai2024sparsellm} and their federated variants~\cite{bai2025fedspallm} represent an emerging direction for enabling FL participants with heterogeneous computational capabilities to collaborate on large model development. These applications share a common need for the cross-HPC facility FL infrastructure our work provides.

\subsection{Practical FL Algorithm Design}
Our systematic comparison reveals that while existing FL algorithms demonstrate the feasibility of cross-facility collaborative training, significant algorithmic innovation remains necessary for production-ready FL on HPC infrastructure. An ideal FL algorithm for geographically distributed HPC environments must address multiple dimensions of heterogeneity simultaneously, including computational capability, queue and scheduling dynamics, and network topology, rather than optimizing for any single factor in isolation.

While FedCompass' success in adapting to computational heterogeneity provides a strong foundation, future algorithms must extend this principle to incorporate queue-aware scheduling. Waiting in a queue effectively removes the client from participation regardless of its processing speed, making queue dynamics as important as computational throughput.

\begin{figure}[!ht]
    \centering
    \includegraphics[width=1.0\linewidth]{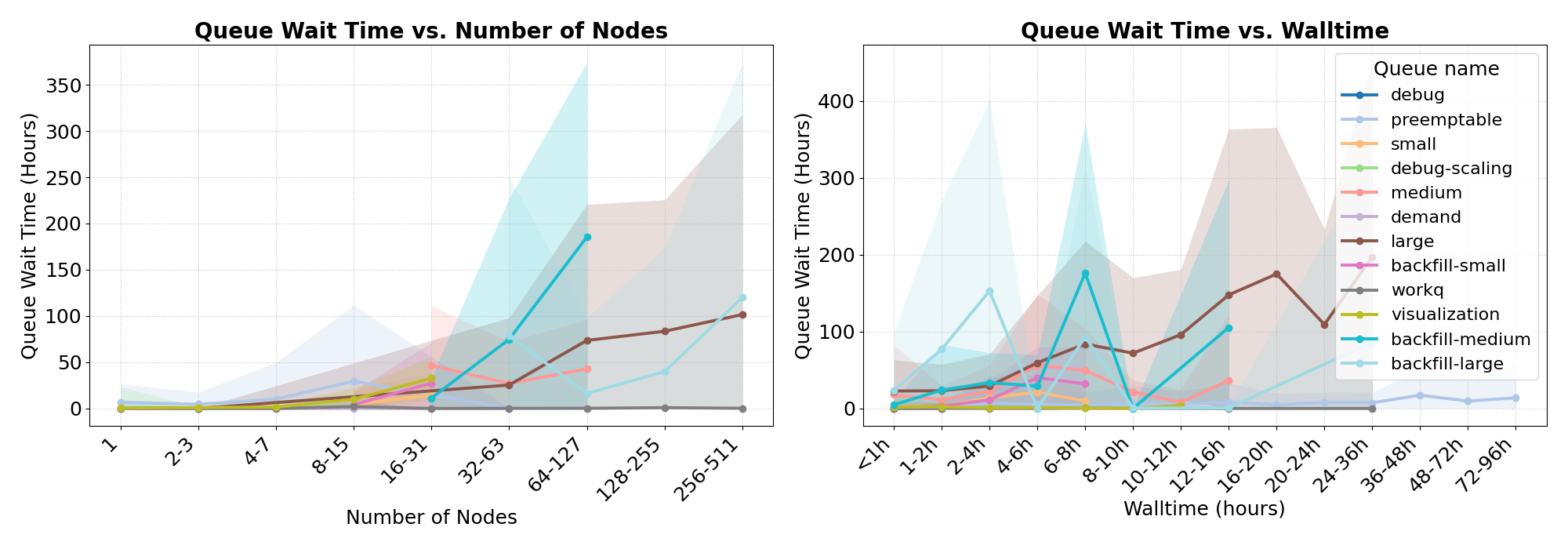}
    \caption{Polaris queue time with varying number of nodes and walltime requested. Solid lines represent the mean queue wait time and shaded regions represent one standard deviation. Different queues are designed to handle jobs with different characteristics in order to maximize supercomputer utilization. Each queue imposes its own constraints on the maximum number of nodes requestable, the maximum walltime allowed, and the allocation of available node-hours.}
    \label{fig:ALCF_queue}
\end{figure}

To investigate the significance of queueing delays, we analyze historical queue data from Polaris obtained from ACLF. Figure~\ref{fig:ALCF_queue} presents these results. Historical analysis shows that large allocations (64+ nodes) face median wait times exceeding 100 hours with high variability, explaining why our large-scale experiments required advance reservations. 

The temporal variability we observed, where transfer speeds and throughput fluctuate with facility workload and network congestion, further necessitates algorithms that maintain running performance estimates and adapt aggregation strategies accordingly. 

The substantial communication cost differences across geographically distributed facilities additionally motivate hierarchical aggregation strategies that exploit geographic locality, aggregating models within co-located facilities before exchanging updates across expensive wide-area links.

Adaptive synchronization that dynamically adjusts between synchronous and bounded-asynchronous modes based on observed heterogeneity could combine FedAvg's convergence quality under favorable conditions with FedAsync's resilience to stragglers. 

Addressing the client drift observed under aggressive local training, where client models diverged more in large-scale configurations, further requires mechanisms such as control variates from SCAFFOLD~\cite{karimireddy2020scaffold} or other drift-correcting updates to maintain global model coherence.

The convergence of these principles, queue awareness, temporal robustness, hierarchical communication, and drift mitigation, defines the frontier for next-generation HPC FL algorithms. These insights align with and extend recent practical experiences in building FL systems~\cite{li2025expbuildingfl}, which similarly emphasize the importance of system-aware algorithm design for real-world deployments.

\subsection{Lessons Learned}
Our deployment across four DOE supercomputers yields several concrete lessons for future cross-facility FL deployments. We organize these around four areas: infrastructure, algorithms, communication and scaling, and operations.

\textbf{Infrastructure.} Globus provides a strong foundation for authentication, authorization, and data transfer, and its broad deployment across research computing facilities means new participants can join a federation without requiring new infrastructure. However, each facility generally requires users to set up their own Globus Compute endpoints and authenticate using facility-specific credentials, and not all facilities currently allow transfer and computation to be triggered by credentials from external institutions. Multi-user Globus Compute endpoints remain limited in deployment. Standardized interfaces for job submission, monitoring, and failure recovery across heterogeneous HPC schedulers exist in various forms but lack broad deployment, generally requiring custom engineering per site.

\textbf{Algorithms.} We observe in our experiments that queue times can affect significantly clients' local trainings and their contributions to the global model, yet no existing FL algorithm explicitly models or adapts to scheduling dynamics. FedCompass' computational adaptation improves over naive approaches, but future algorithms must jointly optimize for compute speed, queue behavior, and network topology. Client drift under non-IID partitioning remains a core challenge, especially in real-world FL across diverse scientific institutions which may encounter extreme heterogeneity, and at large node counts where more local updates amplify local model drift before aggregation.

\textbf{Communication and scaling.} Our results motivate hierarchical aggregation that exploits geographic locality. Larger models improve per-byte transfer efficiency but increase absolute synchronization costs linearly, eventually making frequent aggregation prohibitive without update compression or less frequent synchronization. At very large scales within a single supercomputer, intra-node communication during local training may become a bottleneck, particularly if distributed training frameworks are not optimized for the specific network topology or hardware architecture of the HPC systems.

\textbf{Operational sustainability.} Securing large reservations (64+ nodes) across multiple HPC systems simultaneously required extensive advance coordination with supercomputer operators, a process that does not scale. Smaller queue-based deployments are more accessible but introduce unpredictable delays. Dedicated FL scheduling infrastructure, or a shift to fully asynchronous algorithms that eliminate co-scheduling requirements, will be necessary for production deployments. 

Despite these challenges, our work demonstrates that cross-facility FL with HPC systems is an achievable reality, with clear pathways forward through both algorithmic innovation and systems engineering investment to address the heterogeneity, coordination, and operational complexities inherent in production deployments at scale.

\section{Methods}
\label{sec:methods}

We describe the methods underlying our cross-facility FL deployment, covering the APPFL framework, the DOE supercomputer infrastructure, and the orchestration tools that enable coordinated training across facilities.

\subsection{APPFL and DOE Supercomputers}

APPFL (Advanced Privacy-Preserving Federated Learning) is an open-source software framework whose development is led by researchers from Argonne National Laboratory. APPFL is designed to address heterogeneity and security challenges in FL environments \cite{li2025advances}. The framework features a modular architecture comprising six core technical components: Aggregator, Scheduler, Trainer, Privacy, Communicator, and Compressor, all with user-friendly interfaces for customization and extensibility. APPFL supports both synchronous and asynchronous aggregation strategies, incorporates robust privacy-preserving mechanisms including differential privacy and secure aggregation, and provides comprehensive solutions for various FL paradigms, including vertical, hierarchical, and decentralized configurations. APPFL's extensible design enables researchers to implement, test, and validate novel FL algorithms while facilitating real-world deployment across diverse computational infrastructures, making it an ideal foundation for the heterogeneous, security-conscious environments characteristic of scientific HPC deployments.

The four DOE supercomputers used in this study represent some of the world's most advanced HPC systems, each with distinct architectures and operational policies. ALCF Polaris, located at Argonne National Laboratory, features 560 nodes with each node equipped with one AMD EPYC processor and four NVIDIA A100 GPUs, connected via HPE Slingshot 11 networking and managed by the PBS scheduler. Also at Argonne, ALCF Aurora represents an exascale system with over 9000 nodes, each containing two Intel Xeon Max processors and six Intel Data Center GPU Max 1550 Series accelerators (equivalently 12 GPUs), interconnected through HPE Slingshot 11 networking and employing the PBS scheduler optimized for exascale computing and AI workloads. OLCF Frontier, hosted at Oak Ridge National Laboratory, is an exascale system with over 9400 nodes, each featuring one AMD EPYC processor and four AMD MI250X Graphics Compute Die (eight logical GPUs) connected through HPE Slingshot 11, and operates under a customized SLURM scheduler with complex queue structures designed for diverse scientific workloads. NERSC Perlmutter, based at Lawrence Berkeley National Laboratory, comprises a heterogeneous configuration with over 1500 GPU nodes (each with four NVIDIA A100 GPUs) and over 3000 CPU-only nodes, managed by SLURM with fair share scheduling policies and specialized queues tailored to different workload types. 

These systems differ not only in hardware specifications but also in software environments, job submission interfaces, allocation policies, and operational schedules, creating unique challenges for executing coordinated FL workloads across facilities.

\subsection{Orchestration Tools}

\begin{figure}[!ht]
    \centering
    \includegraphics[width=0.7\linewidth]{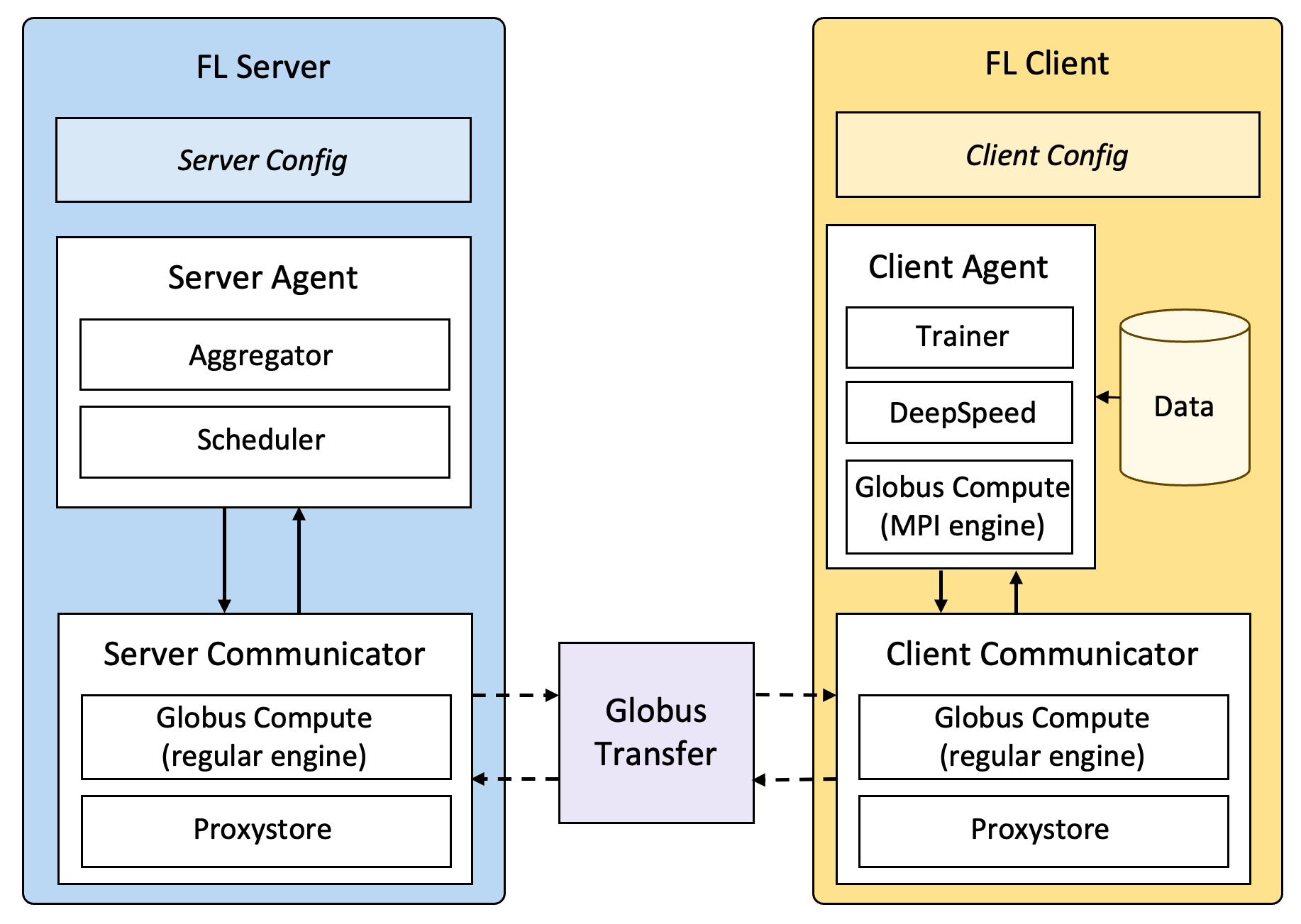}
    \caption{Overall implementation of the cross-facility FL framework. The FL server (blue) comprises a Server Agent handling aggregation and scheduling, and a Server Communicator using Globus Compute and ProxyStore for model exchange. The FL client (yellow) comprises a Client Agent running the Trainer and DeepSpeed via a Globus Compute MPI Engine, and a Client Communicator mirroring the server-side communication components. Model parameters are transferred between server and clients via Globus Transfer.}
    \label{fig:orchetration}
\end{figure}

Figure~\ref{fig:orchetration} shows the overall architecture of our cross-facility framework. We employ Globus Compute \cite{chard2020funcx} to submit local training jobs to distributed HPC clients. As Globus Compute imposes a 10~MB data transfer limit, which is inadequate for transferring model parameters and training metadata of modern machine learning tasks, we integrate Globus Transfer, which provides high-throughput and reliable data movement capabilities. 

More specifically, we employ ProxyStore \cite{pauloski2024object} to orchestrate Globus transfers directly from application code. ProxyStore is a pass-by-reference system that enables transparent object movement across distributed environments by replacing large data objects with lightweight proxies that automatically resolve to the actual data when accessed. This design decision enables automated, seamless integration within our FL workflow while maintaining fine-grained control over data movement timing and destination management. It also enables use of the dedicated data transfer nodes~\cite{dart2013science,chard2018modern} supported by many HPC systems, ensuring that data movement activities do not interfere with computational workloads and providing consistent transfer performance across the distributed infrastructure.

We leverage DeepSpeed, a distributed training framework that provides multiple levels of parallelism essential for modern large-scale machine learning tasks. Specifically, our current implementation uses distributed data parallelism and ZeRO optimization stages, which collectively address two critical computational challenges: efficient utilization of multiple GPU resources for parallel data processing and memory optimization for large model accommodation. The ZeRO optimization stages enable controlled parameter and optimizer state offloading to CPU memory, thereby allowing models such as Llama2-7B to be trained or fine-tuned within the memory constraints of individual GPU devices where they would otherwise be infeasible. Furthermore, adopting DeepSpeed allows for future expansion of parallelism capabilities. 

DeepSpeed's default process launcher, \texttt{pdsh}, is typically unavailable on DOE HPC clusters; we therefore adopt the MPI protocol for parallel job launching, accomplished through the Globus MPI Engine. This engine restricts function execution to MPI and shell-based operations rather than standard Python function calls, which would normally allow direct transfer of function arguments and return values. To overcome this, we introduce a three-engine architecture. A client-side Globus Compute Engine manages local staging of proxied model parameters, training configurations, and metadata from the server, ensuring all training artifacts are available to the MPI Engine before training begins. Following local training, a server-side Globus Compute Engine collects and stages proxied client model parameters and metadata for retrieval during aggregation. Together, these three engines enable efficient coordination of training and aggregation across distributed HPC facilities.

\section{Conclusion}
\label{sec:conclusion}

We have demonstrated the feasibility and characterized the performance of FL across four DOE leadership-class supercomputers, as a representative case of more general cross-facility FL of heterogeneous HPC systems. Our experiments reveal substantial heterogeneity in computational throughput, communication costs, and queueing dynamics across these systems: heterogeneities that existing FL algorithms do not adequately address. Our comparison of algorithms under realistic HPC conditions shows that while computational capability adaptation improves performance, future algorithms must simultaneously account for queue-aware scheduling and temporal variability to achieve practical production deployments. This work establishes both the infrastructure and empirical foundation necessary to enable privacy-preserving collaborative machine learning across geographically distributed scientific institutions, opening pathways for FL applications in domains where data sensitivity, institutional policies, or sheer data scale have previously prevented collaborative model development.

\section*{Acknowledgment}

This work was supported by the U.S. Department of Energy, Office of Science, Advanced Scientific Computing Research, under Contract DE-AC02-06CH11357.
We gratefully acknowledge the computing resources provided on Improv, a high-performance computing cluster operated by the Laboratory Computing Resource Center at Argonne National Laboratory.
An award of computer time was provided by the ASCR Leadership Computing Challenge (ALCC) program. This research used resources of the Argonne Leadership Computing Facility, which is a U.S. Department of Energy Office of Science User Facility operated under contract DE-AC02-06CH11357.
This research used resources of the Oak Ridge Leadership Computing Facility at the Oak Ridge National Laboratory, which is supported by the Office of Science of the U.S. Department of Energy under Contract No. DE-AC05-00OR22725. 
This research used resources of the National Energy Research Scientific Computing Center (NERSC), a Department of Energy User Facility using NERSC award ALCC-ERCAP0038201.
 
\bibliographystyle{plain}
\bibliography{main}

@inproceedings{bonawitz2019federatedlearningscaledesign,
  author = {Bonawitz, Keith and Eichner, Hubert and Grieskamp, Wolfgang and Huba, Dzmitry and Ingerman, Alex and Ivanov, Vladimir and Kiddon, Chloe and Konečný, Jakub and Mazzocchi, Stefano and McMahan, H. Brendan and Van Overveldt, Timon and Petrou, David and Ramage, Daniel and Roselander, Jason},
  title = {Towards Federated Learning at Scale: System Design},
  year = {2019},
  publisher = {MLSys},
  booktitle = {Proceedings of Machine Learning and Systems},
  volume = {1},
  pages = {374--388}
}

@inproceedings{huba2022papayapracticalprivatescalable,
  author={Dzmitry Huba and John Nguyen and Kshitiz Malik and Ruiyu Zhu and Mike Rabbat and Ashkan Yousefpour and Carole-Jean Wu and Hongyuan Zhan and Pavel Ustinov and Harish Srinivas and Kaikai Wang and Anthony Shoumikhin and Jesik Min and Mani Malek},
  title = {Papaya: Practical, Private, and Scalable Federated Learning},
  year = {2022},
  publisher = {MLSys},
  booktitle = {Proceedings of Machine Learning and Systems},
  volume = {4},
  pages = {922--939}
}

@article{paulik2021federatedevaluationtuningondevice,
      title={Federated Evaluation and Tuning for On-Device Personalization: System Design \& Applications}, 
      author={Matthias Paulik and Matt Seigel and Henry Mason and Dominic Telaar and Joris Kluivers and Rogier van Dalen and Chi Wai Lau and Luke Carlson and Filip Granqvist and Chris Vandevelde and Sudeep Agarwal and Julien Freudiger and Andrew Byde and Abhishek Bhowmick and Gaurav Kapoor and Si Beaumont and Áine Cahill and Dominic Hughes and Omid Javidbakht and Fei Dong and Rehan Rishi and Stanley Hung},
      year={2021},
      eprint={2102.08503},
      archivePrefix={arXiv},
      primaryClass={cs.LG},
      url={https://arxiv.org/abs/2102.08503},
      journal={Preprint arXiv:2102.08503}
      
}

@inproceedings{gainaru2019schedulingscientificworkflows,
  author = {Gainaru, Ana and Aupy, Guillaume and Benoit, Anne and Cappello, Franck and Robert, Yves and Snir, Marc},
  title = {Scheduling the {I/O of HPC} Applications Under Congestion},
  booktitle = {IEEE International Parallel and Distributed Processing Symposium},
  year = {2015},
  pages = {1013--1022}
}

@inproceedings{mcmahan2017communication,
  title={Communication-efficient learning of deep networks from decentralized data},
  author={McMahan, Brendan and Moore, Eider and Ramage, Daniel and Hampson, Seth and y Arcas, Blaise Aguera},
  booktitle = {Proceedings of the Twentieth International Conference on Artificial Intelligence and Statistics},
  series    = {Proceedings of Machine Learning Research},
  volume    = {54},
  year = {2017},
  publisher = {PMLR},
}

@article{kairouz2021advances,
  title={Advances and open problems in federated learning},
  author={Peter Kairouz and H. Brendan McMahan and Brendan Avent and Aurélien Bellet and Mehdi Bennis and Arjun Nitin Bhagoji and Kallista Bonawitz and Zachary Charles and Graham Cormode and Rachel Cummings and Rafael G. L. D'Oliveira and Hubert Eichner and Salim El Rouayheb and David Evans and Josh Gardner and Zachary Garrett and Adrià Gascón and Badih Ghazi and Phillip B. Gibbons and Marco Gruteser and Zaid Harchaoui and Chaoyang He and Lie He and Zhouyuan Huo and Ben Hutchinson and Justin Hsu and Martin Jaggi and Tara Javidi and Gauri Joshi and Mikhail Khodak and Jakub Konečný and Aleksandra Korolova and Farinaz Koushanfar and Sanmi Koyejo and Tancrède Lepoint and Yang Liu and Prateek Mittal and Mehryar Mohri and Richard Nock and Ayfer Özgür and Rasmus Pagh and Mariana Raykova and Hang Qi and Daniel Ramage and Ramesh Raskar and Dawn Song and Weikang Song and Sebastian U. Stich and Ziteng Sun and Ananda Theertha Suresh and Florian Tramèr and Praneeth Vepakomma and Jianyu Wang and Li Xiong and Zheng Xu and Qiang Yang and Felix X. Yu and Han Yu and Sen Zhao},
  journal={Foundations and trends{\textregistered} in machine learning},
  volume={14},
  number={1--2},
  pages={1--210},
  year={2021},
  publisher={Now Publishers, Inc.}
}

@article{frey2002condor,
  title={Condor-{G}: A computation management agent for multi-institutional grids},
  author={Frey, James and Tannenbaum, Todd and Livny, Miron and Foster, Ian and Tuecke, Steven},
  journal={Cluster Computing},
  volume={5},
  number={3},
  pages={237--246},
  year={2002},
  publisher={Springer}
}

@inproceedings{li2024fedcompass,
  title={Fed{C}ompass: Efficient cross-silo federated learning on heterogeneous client devices using a computing power aware scheduler},
  author={Li, Zilinghan and Chaturvedi, Pranshu and He, Shilan and Chen, Han and Singh, Gagandeep and Kindratenko, Volodymyr and Huerta, Eliu A and Kim, Kibaek and Madduri, Ravi},
  booktitle={The International Conference on Learning Representations},
  year={2024}
}

@inproceedings{ryu2022appfl,
  title = {{APPFL}: Open-Source Software Framework for Privacy-Preserving Federated Learning},
  url = {http://dx.doi.org/10.1109/IPDPSW55747.2022.00175},
  DOI = {10.1109/ipdpsw55747.2022.00175},
  booktitle = {IEEE International Parallel and Distributed Processing Symposium Workshops (IPDPSW)},
  publisher = {IEEE},
  author = {Ryu,  Minseok and Kim,  Youngdae and Kim,  Kibaek and Madduri,  Ravi K.},
  year = {2022},
  pages = {1074-1083},
  month = may 
}

@inproceedings{li2025advances,
  title = {Advances in {APPFL}: a Comprehensive and Extensible Federated Learning Framework},
  url = {http://dx.doi.org/10.1109/CCGRID64434.2025.00031},
  DOI = {10.1109/ccgrid64434.2025.00031},
  booktitle = {IEEE 25th International Symposium on Cluster,  Cloud and Internet Computing (CCGrid)},
  publisher = {IEEE},
  author = {Li,  Zilinghan and He, Shilan and Yang, Ze and Ryu, Minseok and Kim, Kibaek and Madduri, Ravi},
  year = {2025},
  pages = {103-113},
  month = may
}

@article{liu2021fate,
  title={{FATE}: An industrial grade platform for collaborative learning with data protection},
  author={Liu, Yang and Fan, Tao and Chen, Tianjian and Xu, Qian and Yang, Qiang},
  journal={Journal of Machine Learning Research},
  volume={22},
  number={226},
  pages={1--6},
  year={2021}
}

@article{federatedscope,
  title = {Federated{S}cope: A Flexible Federated Learning Platform for Heterogeneity},
  author = {Xie, Yuexiang and Wang, Zhen and Gao, Dawei and Chen, Daoyuan and Yao, Liuyi and Kuang, Weirui and Li, Yaliang and Ding, Bolin and Zhou, Jingren},
  journal={Proceedings of the VLDB Endowment},
  volume={16},
  number={5},
  pages={1059--1072},
  year={2023}
}

@article{roth2022nvidia,
  doi = {10.48550/ARXIV.2210.13291},
  url = {https://arxiv.org/abs/2210.13291},
  author = {Roth, Holger R. and Cheng, Yan and Wen, Yuhong and Yang, Isaac and Xu, Ziyue and Hsieh, Yuan-Ting and Kersten, Kristopher and Harouni, Ahmed and Zhao, Can and Lu, Kevin and Zhang, Zhihong and Li, Wenqi and Myronenko, Andriy and Yang, Dong and Yang, Sean and Rieke, Nicola and Quraini, Abood and Chen, Chester and Xu, Daguang and Ma, Nic and Dogra, Prerna and Flores, Mona and Feng, Andrew},
  title = {{NVIDIA FLARE}: Federated Learning from Simulation to Real-World},
  publisher = {arXiv},
  year = {2022},
  journal={Preprint arXiv:2210.13291},
  copyright = {arXiv.org perpetual, non-exclusive license}
}

@article{reina2021openfl,
   title={OpenFL: the open federated learning library},
   volume={67},
   ISSN={1361-6560},
   url={http://dx.doi.org/10.1088/1361-6560/ac97d9},
   DOI={10.1088/1361-6560/ac97d9},
   number={21},
   journal={Physics in Medicine \&amp; Biology},
   publisher={IOP Publishing},
   author={Foley, Patrick and Sheller, Micah J and Edwards, Brandon and Pati, Sarthak and Riviera, Walter and Sharma, Mansi and Narayana Moorthy, Prakash and Wang, Shih-han and Martin, Jason and Mirhaji, Parsa and Shah, Prashant and Bakas, Spyridon},
   year={2022},
   month=oct, pages={214001} }

@inproceedings{chard2020funcx,
  title={Func{X}: A federated function serving fabric for science},
  author={Chard, Ryan and Babuji, Yadu and Li, Zhuozhao and Skluzacek, Tyler and Woodard, Anna and Blaiszik, Ben and Foster, Ian and Chard, Kyle},
  booktitle={Proceedings of the 29th International symposium on high-performance parallel and distributed computing},
  pages={65--76},
  year={2020}
}

@article{hudson2025flight,
  title={Flight: A {FaaS}-based framework for complex and hierarchical federated learning},
  author={Hudson, Nathaniel and Hayot-Sasson, Valerie and Babuji, Yadu and Baughman, Matt and Pauloski, J Gregory and Chard, Ryan and Foster, Ian and Chard, Kyle},
  journal={Future Generation Computer Systems},
  pages={107998},
  volume={174},
  year={2026},
  publisher={Elsevier}
}

@article{pauloski2024object,
  title={Object proxy patterns for accelerating distributed applications},
  author={Pauloski, J Gregory and Hayot-Sasson, Valerie and Ward, Logan and Brace, Alexander and Bauer, Andr{\'e} and Chard, Kyle and Foster, Ian},
  journal={IEEE Transactions on Parallel and Distributed Systems},
  year={2025},
  volume={36},
  number={2},
  pages={253-265},
  publisher={IEEE}
}

@inproceedings{kim2024privacy,
  title = {Privacy-Preserving Federated Learning for Science: Challenges and Research Directions},
  url = {http://dx.doi.org/10.1109/BigData62323.2024.10825853},
  DOI = {10.1109/bigdata62323.2024.10825853},
  booktitle = {2024 IEEE International Conference on Big Data (BigData)},
  publisher = {IEEE},
  author = {Kim,  Kibaek and Raghavan,  Krishnan and Kotevska,  Olivera and Dorier,  Matthieu and Madduri,  Ravi and Ryu,  Minseok and Munson,  Todd and Ross,  Rob and Flynn,  Thomas and Kagawa,  Ai and Yoon,  Byung-Jun and Engelmann,  Christian and Yousefian,  Farzad},
  year = {2024},
  month = dec,
  pages = {7849–7853}
}

@article{10943420241281744,
author = {Weijian Zheng and Jack Kordas and Tyler J Skluzacek and Raj Kettimuthu and Ian Foster},
title ={Globus service enhancements for exascale applications and facilities},
journal = {The International Journal of High Performance Computing Applications},
volume = {38},
number = {6},
pages = {658-670},
year = {2024},
doi = {10.1177/10943420241281744},
URL = { 
https://doi.org/10.1177/10943420241281744
}
}

@inproceedings{czajkowski2002snap,
  title={{SNAP}: A protocol for negotiating service level agreements and coordinating resource management in distributed systems},
  author={Czajkowski, Karl and Foster, Ian and Kesselman, Carl and Sander, Volker and Tuecke, Steven},
  booktitle={Workshop on Job Scheduling Strategies for Parallel Processing},
  pages={153--183},
  year={2002},
  organization={Springer}
}

@inproceedings{yu2024llasmol,
  title={Lla{SM}ol: Advancing Large Language Models for Chemistry with a Large-Scale, Comprehensive, High-Quality Instruction Tuning Dataset},
  author={Botao Yu and Frazier N. Baker and Ziqi Chen and Xia Ning and Huan Sun},
  booktitle={First Conference on Language Modeling},
  year={2024},
  url={https://openreview.net/forum?id=lY6XTF9tPv}
}

@article{li2025expbuildingfl,
      title={Experiences Building Enterprise-Level Privacy-Preserving Federated Learning to Power {AI} for Science}, 
      author={Zilinghan Li and Aditya Sinha and Yijiang Li and Kyle Chard and Kibaek Kim and Ravi Madduri},
      year={2025},
      eprint={2511.08998},
      archivePrefix={arXiv},
      primaryClass={cs.DC},
      url={https://arxiv.org/abs/2511.08998},
      journal={Preprint arXiv:2511.08998}
}

@article{beutel2020flower,
      title={Flower: A Friendly Federated Learning Research Framework}, 
      author={Daniel J. Beutel and Taner Topal and Akhil Mathur and Xinchi Qiu and Javier Fernandez-Marques and Yan Gao and Lorenzo Sani and Kwing Hei Li and Titouan Parcollet and Pedro Porto Buarque de Gusmão and Nicholas D. Lane},
      year={2022},
      eprint={2007.14390},
      archivePrefix={arXiv},
      primaryClass={cs.LG},
      url={https://arxiv.org/abs/2007.14390}, 
      journal={Preprint arXiv:2007.14390}
}

@inproceedings{li2020federatedoptimization,
  title={Federated Optimization in Heterogeneous Networks},
  author={Li, Tian and Sahu, Anit Kumar and Zaheer, Manzil and Sanjabi, Maziar and Talwalkar, Ameet and Smith, Virginia},
  booktitle={Proceedings of Machine Learning and Systems},
  volume={2},
  pages={429--450},
  year={2020}
}

@inproceedings{karimireddy2020scaffold,
  title={{SCAFFOLD}: Stochastic Controlled Averaging for Federated Learning},
  author={Karimireddy, Sai Praneeth and Kale, Satyen and Mohri, Mehryar and Reddi, Sashank and Stich, Sebastian and Suresh, Ananda Theertha},
  booktitle={International Conference on Machine Learning},
  pages={5132--5143},
  year={2020},
  organization={PMLR}
}

@inproceedings{reddi2020adaptive,
  title={Adaptive Federated Optimization},
  author={Reddi, Sashank and Charles, Zachary and Zaheer, Manzil and Garrett, Zachary and Rush, Keith and Kone{\v{c}}n{\`y}, Jakub and Kumar, Sanjiv and McMahan, H Brendan},
  booktitle={International Conference on Learning Representations},
  year={2021}
}

@article{xie2019asynchronousfederatedoptimization,
      title={Asynchronous Federated Optimization}, 
      author={Cong Xie and Sanmi Koyejo and Indranil Gupta},
      year={2020},
      eprint={1903.03934},
      archivePrefix={arXiv},
      primaryClass={cs.DC},
      url={https://arxiv.org/abs/1903.03934},
      journal={Preprint arXiv:1903.03934},
}

@inproceedings{nguyen2022fedbufffederatedlearning,
  title={{FedBuff}: Federated Learning with Buffered Asynchronous Aggregation},
  author={Nguyen, John and Malik, Kshitiz and Zhan, Hongyuan and Yousefpour, Ashkan and Rabbat, Mike and Malek, Mani and Huba, Dzmitry},
  booktitle = {Proceedings of the Twenty-fifth International Conference on Artificial Intelligence and Statistics},
  series    = {Proceedings of Machine Learning Research},
  volume    = {151},
  year      = {2022},
  publisher = {PMLR}
}

@article{huang2022cross,
  title={Cross-silo federated learning: Challenges and opportunities},
  author={Huang, Chao and Huang, Jianwei and Liu, Xin},
  journal={Preprint arXiv:2206.12949},
  year={2022}
}

@inproceedings{liu2020clientedge,
  title={Client-Edge-Cloud Hierarchical Federated Learning},
  author={Liu, Lumin and Zhang, Jun and Song, S H and Letaief, Khaled B},
  booktitle={IEEE International Conference on Communications},
  pages={1--6},
  year={2020},
  organization={IEEE}
}

@inproceedings{yoo2003slurm,
  title={{SLURM}: Simple {L}inux Utility for Resource Management},
  author={Yoo, Andy B. and Jette, Morris A. and Grondona, Mark},
  booktitle={Workshop on Job Scheduling Strategies for Parallel Processing},
  pages={44--60},
  year={2003},
  organization={Springer}
}

@inproceedings{cobalt2010,
  title={Cobalt: An open source platform for {HPC} system software research},
  author={Desai, Narayan},
  booktitle={Edinburgh BG/L System Software Workshop},
  pages={803--820},
  note={\url{https://ftp.mcs.anl.gov/pub/cobalt/papers/cobalt-epcc-10-05.pdf}},
  year={2005}
}

@article{deelman2015pegasus,
  title={Pegasus: A Workflow Management System for Science Automation},
  author={Deelman, Ewa and Vahi, Karan and Juve, Gideon and Rynge, Mats and Callaghan, Scott and Maechling, Philip J and Mayani, Rajiv and Chen, Weiwei and Da Silva, Rafael Ferreira and Livny, Miron and Wenger, Kent},
  journal={Future Generation Computer Systems},
  volume={46},
  pages={17--35},
  year={2015},
  publisher={Elsevier}
}

@article{hoefler2017performance,
  title={Scientific Benchmarking of Parallel Computing Systems: Twelve Ways to Tell the Masses when Reporting Performance Results},
  author={Hoefler, Torsten and Belli, Roberto},
  journal={SC'15: International Conference for High Performance Computing, Networking, Storage and Analysis},
  year={2015},
  publisher={ACM}
}

@article{rasley2020deepspeed,
  title={{DeepSpeed}: System Optimizations Enable Training Deep Learning Models with Over 100 Billion Parameters},
  author={Rasley, Jeff and Rajbhandari, Samyam and Ruwase, Olatunji and He, Yuxiong},
  journal={Proceedings of the 26th ACM SIGKDD International Conference on Knowledge Discovery \& Data Mining},
  pages={3505--3506},
  year={2020}
}

@article{rajbhandari2020zero,
  title={{ZeRO}: Memory Optimizations Toward Training Trillion Parameter Models},
  author={Rajbhandari, Samyam and Rasley, Jeff and Ruwase, Olatunji and He, Yuxiong},
  journal={SC20: International Conference for High Performance Computing, Networking, Storage and Analysis},
  pages={1--16},
  year={2020},
  organization={IEEE}
}

@article{sergeev2018horovod,
  title={Horovod: Fast and Easy Distributed Deep Learning in {TensorFlow}},
  author={Sergeev, Alexander and Del Balso, Mike},
  journal={Preprint arXiv:1802.05799},
  year={2018}
}

@inproceedings{chen2018powersgd,
  title={{PowerSGD}: Practical Low-Rank Gradient Compression for Distributed Optimization},
  author={Vogels, Thijs and Karimireddy, Sai Praneeth and Jaggi, Martin},
  booktitle={Advances in Neural Information Processing Systems},
  volume={32},
  year={2019}
}

@inproceedings{bai2024sparsellm,
  title={Sparse{LLM}: Towards Global Pruning of Pre-trained Language Models},
  author={Bai, Guangji and Li, Yijiang and Ling, Chen and Kim, Kibaek and Zhao, Liang},
  booktitle={Advances in Neural Information Processing Systems},
  volume={38},
  year={2024}
}

@inproceedings{bai2025fedspallm,
  title={{FedSpaLLM}: Federated Pruning of Large Language Models},
  author={Bai, Guangji and Li, Yijiang and Li, Zilinghan and Kim, Kibaek and Zhao, Liang},
  booktitle={Conference of the Nations of the Americas Chapter of the Association for Computational Linguistics:},
  volume={1},
  year={2025}
}

@article{rieke2020future,
  title={The Future of Digital Health with Federated Learning},
  author={Rieke, Nicola and Hancox, Jonny and Li, Wenqi and Milletari, Fausto and Roth, Holger R. and Albarqouni, Shadi and Bakas, Spyridon and Galtier, Mathieu N and Landman, Bennett A and Maier-Hein, Klaus and Ourselin, Sebastien and Sheller, Micah and Summers, Ronald M. and Trask, Andrew and Xu, Daguang and Baust, Maximilian and Cardoso, M. Jorge},
  journal={NPJ Digital Medicine},
  volume={3},
  number={119},
  pages={},
  year={2020},
  publisher={Nature Publishing Group}
}

@article{sheller2020federated,
  title={Federated Learning in Medicine: Facilitating Multi-Institutional Collaborations Without Sharing Patient Data},
  author={Sheller, Micah J and Edwards, Brandon and Reina, G Anthony and Martin, Jason and Pati, Sarthak and Kotrotsou, Aikaterini and Milchenko, Mikhail and Xu, Weilin and Marcus, Daniel and Colen, Rivka R and Bakas, Spyridon},
  journal={Scientific Reports},
  volume={10},
  number={12598},
  pages={},
  year={2020},
  publisher={Nature Publishing Group}
}

@article{melloddy2022collaborative,
  title={{MELLODDY}: Cross-pharma Federated Learning at Unprecedented Scale Unlocks Benefits in QSAR without Compromising Proprietary Information},
  author={Heyndrickx, Wouter and Mervin, Lewis and Morawietz, Tobias 
          and Sturm, No\'{e} and Friedrich, Lukas and Zalewski, Adam 
          and Pentina, Anastasia and Humbeck, Lina and Oldenhof, Martijn 
          and Niwayama, Ritsuya and Schmidtke, Peter and Fechner, Nikolas 
          and Simm, Jaak and Arany, Adam and Drizard, Nicolas 
          and Jabal, Rama and Afanasyeva, Arina and Loeb, Regis 
          and Verma, Shlok and Harnqvist, Simon and Holmes, Matthew 
          and Pejo, Balazs and Telenczuk, Maria and Holway, Nicholas 
          and Dieckmann, Arne and Rieke, Nicola and Zumsande, Friederike 
          and Clevert, Djork-Arn\'{e} and Krug, Michael and Luscombe, Christopher 
          and Green, Darren and Ertl, Peter and Antal, Peter 
          and Marcus, David and Do Huu, Nicolas and Fuji, Hideyoshi 
          and Pickett, Stephen and Acs, Gergely and Boniface, Eric 
          and Beck, Bernd and Sun, Yax and Gohier, Arnaud 
          and Rippmann, Friedrich and Engkvist, Ola and G\"{o}ller, Andreas H 
          and Moreau, Yves and Galtier, Mathieu N and Schuffenhauer, Ansgar 
          and Ceulemans, Hugo},
  journal={Journal of chemical information and modeling},
  volume={64},
  number={7},
  pages={2331--2344},
  year={2024},
}

@article{bose2024rnnsfoundationmodelsempirical,
      title={From {RNNs} to Foundation Models: An Empirical Study on Commercial Building Energy Consumption}, 
      author={Shourya Bose and Yijiang Li and Amy Van Sant and Yu Zhang and Kibaek Kim},
      year={2024},
      eprint={2411.14421},
      archivePrefix={arXiv},
      primaryClass={cs.LG},
      url={https://arxiv.org/abs/2411.14421}, 
      journal={Preprint arXiv:2411.14421}
}

@article{bose2023privacypreservingloadforecastingpersonalized,
      title={Privacy-Preserving Load Forecasting via Personalized Model Obfuscation}, 
      author={Shourya Bose and Yu Zhang and Kibaek Kim},
      year={2023},
      eprint={2312.00036},
      archivePrefix={arXiv},
      primaryClass={cs.CR},
      url={https://arxiv.org/abs/2312.00036},
      journal={Preprint arXiv:2312.00036}
}

@InProceedings{10.1007/3-540-60153-8_34,
    author="Henderson, Robert L.",
    title="Job scheduling under the {Portable Batch System}",
    booktitle="Job Scheduling Strategies for Parallel Processing",
    year="1995",
    publisher="Springer Berlin Heidelberg",
    address="Berlin, Heidelberg",
    pages="279--294",
}

@article{touvron2023llama2openfoundation,
      title={Llama 2: Open Foundation and Fine-Tuned Chat Models}, 
      author={Hugo Touvron and Louis Martin and Kevin Stone and Peter Albert and Amjad Almahairi and Yasmine Babaei and Nikolay Bashlykov and Soumya Batra and Prajjwal Bhargava and Shruti Bhosale and Dan Bikel and Lukas Blecher and Cristian Canton Ferrer and Moya Chen and Guillem Cucurull and David Esiobu and Jude Fernandes and Jeremy Fu and Wenyin Fu and Brian Fuller and Cynthia Gao and Vedanuj Goswami and Naman Goyal and Anthony Hartshorn and Saghar Hosseini and Rui Hou and Hakan Inan and Marcin Kardas and Viktor Kerkez and Madian Khabsa and Isabel Kloumann and Artem Korenev and Punit Singh Koura and Marie-Anne Lachaux and Thibaut Lavril and Jenya Lee and Diana Liskovich and Yinghai Lu and Yuning Mao and Xavier Martinet and Todor Mihaylov and Pushkar Mishra and Igor Molybog and Yixin Nie and Andrew Poulton and Jeremy Reizenstein and Rashi Rungta and Kalyan Saladi and Alan Schelten and Ruan Silva and Eric Michael Smith and Ranjan Subramanian and Xiaoqing Ellen Tan and Binh Tang and Ross Taylor and Adina Williams and Jian Xiang Kuan and Puxin Xu and Zheng Yan and Iliyan Zarov and Yuchen Zhang and Angela Fan and Melanie Kambadur and Sharan Narang and Aurelien Rodriguez and Robert Stojnic and Sergey Edunov and Thomas Scialom},
      year={2023},
      eprint={2307.09288},
      archivePrefix={arXiv},
      primaryClass={cs.CL},
      url={https://arxiv.org/abs/2307.09288}, 
      journal={Preprint arXiv:2307.09288}
}

@misc{aws-ofi-nccl,
  author = {{Amazon Web Services}},
  title = {{AWS OFI NCCL}: {AWS} Libfabric Plugin for {NCCL} and {RCCL}},
  year = {2025},
  url = {https://github.com/aws/aws-ofi-nccl},
  note = {Accessed: 2026-02-19}
}

@inproceedings{li2020convergencefedavgnoniiddata,
  author    = {Xiang Li and Kaixuan Huang and Wenhao Yang and Shusen Wang and Zhihua Zhang},
  title     = {On the Convergence of FedAvg on Non-IID Data},
  booktitle = {International Conference on Learning Representations},
  year      = {2020},
  url       = {https://openreview.net/forum?id=HJxNAnVtDS}
}

@article{chard2018modern,
  title={The {Modern Research Data Portal}: A design pattern for networked, data-intensive science},
  author={Chard, Kyle and Dart, Eli and Foster, Ian and Shifflett, David and Tuecke, Steven and Williams, Jason},
  journal={PeerJ Computer Science},
  volume={4},
  pages={e144},
  year={2018},
  publisher={PeerJ Inc.}
}

@inproceedings{dart2013science,
  title={The {Science DMZ}: A network design pattern for data-intensive science},
  author={Dart, Eli and Rotman, Lauren and Tierney, Brian and Hester, Mary and Zurawski, Jason},
  booktitle={Proceedings of the International Conference on High Performance Computing, Networking, Storage and Analysis},
  pages={1--10},
  year={2013}
}

@article{li2026luminafoundationmodelstopology,
      title={{LUMINA}: Foundation Models for Topology Transferable {ACOPF}}, 
      author={Yijiang Li and Zeeshan Memon and Hongwei Jin and Stefano Fenu and Keunju Song and Sunash B Sharma and Parfait Gasana and Hongseok Kim and Liang Zhao and Kibaek Kim},
      year={2026},
      eprint={2603.04300},
      archivePrefix={arXiv},
      primaryClass={cs.LG},
      url={https://arxiv.org/abs/2603.04300}, 
      journal={Preprint arXiv:2603.04300}
}

@article{he2020fedml,
  title={{FedML}: A research library and benchmark for federated machine learning},
  author={He, Chaoyang and Li, Songze and So, Jinhyun and Zeng, Xiao and Zhang, Mi and Wang, Hongyi and Wang, Xiaoyang and Vepakomma, Praneeth and Singh, Abhishek and Qiu, Hang and others},
  journal={arXiv preprint arXiv:2007.13518},
  year={2020}
}

@inproceedings{chai2021fedat,
  title={{FedAT}: A high-performance and communication-efficient federated learning system with asynchronous tiers},
  author={Chai, Zheng and Chen, Yujing and Anwar, Ali and Zhao, Liang and Cheng, Yue and Rangwala, Huzefa},
  booktitle={Proceedings of the international conference for high performance computing, networking, storage and analysis},
  pages={1--16},
  year={2021}
}

@article{colonnelli2024cross,
  title={Cross-facility federated learning},
  author={Colonnelli, Iacopo and Birke, Robert and Malenza, Giulio and Mittone, Gianluca and Mulone, Alberto and Galjaard, Jeroen and Chen, Lydia Y and Bassini, Sanzio and Scipione, Gabriella and Martinovi{\v{c}}, Jan and others},
  journal={Procedia Computer Science},
  volume={240},
  pages={3--12},
  year={2024},
  publisher={Elsevier}
}

@article{li2024secure,
  title={Secure Federated Learning Across Heterogeneous Cloud and High-Performance Computing Resources: A Case Study on Federated Fine-Tuning of {LLaMA} 2},
  author={Li, Zilinghan and He, Shilan and Chaturvedi, Pranshu and Kindratenko, Volodymyr and Huerta, Eliu A and Kim, Kibaek and Madduri, Ravi},
  journal={Computing in Science \& Engineering},
  volume={26},
  number={3},
  pages={52--58},
  year={2024},
  publisher={IEEE}
}

@article{zhang2025fedcspc,
  title={{FedCSpc}: A cross-silo federated learning system with error-bounded lossy parameter compression},
  author={Zhang, Zhaorui and Di, Sheng and Zhao, Kai and Jin, Sian and Tao, Dingwen and Ji, Zhuoran and Liu, Benben and Alharthi, Khalid Ayed and Cao, Jiannong and Cappello, Franck},
  journal={IEEE Transactions on Parallel and Distributed Systems},
  year={2025},
  publisher={IEEE}
}

% \section*{Data Availability}

% \section*{Code Availability}

\begin{mdframed}
The submitted manuscript has been created by UChicago Argonne, LLC, Operator of Argonne National Laboratory (``Argonne”). Argonne, a U.S. Department of Energy Office of Science laboratory, is operated under Contract No. DE-AC02-06CH11357. The U.S. Government retains for itself, and others acting on its behalf, a paid-up nonexclusive, irrevocable worldwide license in said article to reproduce, prepare derivative works, distribute copies to the public, and perform publicly and display publicly, by or on behalf of the Government. The Department of Energy will provide public access to these results of federally sponsored research in accordance with the DOE Public Access Plan (http://energy.gov/downloads/doe-public-access-plan).
\end{mdframed}

\end{document}